\documentclass[journal]{IEEEtran}
\usepackage{mathtools}
\usepackage{amscd}
\usepackage{amsthm,amssymb}
\usepackage{amsmath}
\usepackage{cite}
\usepackage[utf8]{inputenc} 
\usepackage{calc,ifthen,moreverb}
\usepackage{graphicx,color,psfrag}
\usepackage{lscape,pdflscape}
\usepackage{multicol}
\usepackage[mathcal]{euscript}
\usepackage{url}
\usepackage{soul}
\usepackage[caption=false,font=footnotesize]{subfig}
\usepackage{bm}

\newcommand{\norm}[1]{\left\lVert #1 \right\rVert} %A new command for the norm symbol
\theoremstyle{definition} % Command to call definition
\theoremstyle{remark} % Command to call definition

\newcommand{\eq}{\begin{equation}}
\newcommand{\qe}{\end{equation}}
\newcommand{\eqarray}{\begin{array}{ll}}
\newcommand{\qearray}{\end{array}}

\def\*#1{\bm{#1}} %for boldface symbols
\def\~#1{\widetilde{#1}} %for tilde
\def\b~#1{\bm{\widetilde{#1}}} %for tilde and boldface
\def\expec{\bm{\text{E}}} %for tilde and boldface

\newcounter{tempEquationCounter} 
\newcounter{thisEquationNumber}

\begin{document}
\title{Geometric-Algebra LMS Adaptive Filter and its Application to Rotation Estimation}

\author{Wilder~B.~Lopes,
        Anas~Al-Nuaimi,
        Cassio~G.~Lopes% <-this % stops a space
\thanks{W.~B.~Lopes (wilder@usp.br) and C.~G.~Lopes (cassio@lps.usp.br) are with the Department of Electronic Systems Engineering, University of Sao Paulo, Brazil. A.~Al-Nuaimi (anas.alnuaimi@tum.de) is with the Chair of Media Technology (LMT), TU München, Germany. The first author was supported by CAPES Foundation, Ministry of Education of Brazil, under Grant BEX 14601/13-3. This work was conducted during a research stay at LMT - TU München.}% 
}%


\markboth{}%
{Shell \MakeLowercase{\textit{et al.}}: Bare Demo of IEEEtran.cls for Journals}
%\ifCLASSOPTIONpeerreview

\maketitle

\begin{abstract}
This paper exploits Geometric (Clifford) Algebra (GA) theory in order to devise and introduce a new adaptive filtering strategy. From a least-squares cost function, the gradient is calculated following results from Geometric Calculus (GC), the extension of GA to handle differential and integral calculus. The novel GA least-mean-squares (GA-LMS) adaptive filter, which inherits properties from standard adaptive filters and from GA, is developed to recursively estimate a rotor (multivector), a hypercomplex quantity able to describe rotations in any dimension. The adaptive filter (AF) performance is assessed via a 3D point-clouds registration problem, which contains a rotation estimation step. Calculating the AF computational complexity suggests that it can contribute to reduce the cost of a full-blown 3D registration algorithm, especially when the number of points to be processed grows. Moreover, the employed GA/GC framework allows for easily applying the resulting filter to estimating rotors in higher dimensions. 
\end{abstract}


%% For peer review papers, you can put extra information on the cover
%% page as needed:
%\ifCLASSOPTIONpeerreview
%\begin{center} \bfseries EDICS Category: SAS-ADAP \end{center}
%\fi
%%
%% For peerreview papers, this IEEEtran command inserts a page break and
%% creates the second title. It will be ignored for other modes.
%\IEEEpeerreviewmaketitle

\section{Introduction}
\label{sec:intro}

\IEEEPARstart{S}{tandard} adaptive filtering theory is based on vector calculus and matrix/linear algebra. Given a cost function, usually a least-(mean)-squares criterion, gradient-descent methods are employed, resulting in a myriad of adaptive algorithms that minimize the original cost function in an adaptive manner~\cite{Diniz2013,Sayed08}. 

This work introduces a new adaptive filtering technique based on GA and GC. Such frameworks generalize linear algebra and vector calculus for hypercomplex variables, specially regarding the representation of geometric transformations~\cite{Hitzer_Introduction,hestenes1999newfoundations,hestenes1987GAcalculus,
Dorst2007GACV,hitzer2013outerRotations,doran2003geometric,DoranDissertation}. In this sense, the GA-LMS is devised in light of GA and using results from GC (instead of vector calculus). The new approach is motivated via an actual computer vision problem, namely 3D registration of point clouds~\cite{Rusu_ICRA2011_PCL}. To validate the algorithm, simulations are run in artificial and real data. The new GA adaptive filtering technique renders an algorithm that may ultimately be a candidate for real-time online rotation estimation. 

\section{Standard Rotation Estimation}
\label{sec:3D_reg_prom}

Consider two sets of points -- point clouds (PCDs) -- in the $\mathbb{R}^3$, Y (\textit{Target}) and X (\textit{Source}), related via a 1-1 correspondence, in which X is a rotated version of Y. Each PCD has $K$ points, $\lbrace y_n \rbrace \in$~Y and $\lbrace x_n \rbrace \in$~X, $n=1...K$, and their centroids are located at the coordinate system origin. 

In the registration process, one needs to find the \emph{linear operator}, i.e., the $3{\times}3$ rotation matrix $R$ (\hspace{-0.1mm}\cite{Meyer01}, p.320), that maps X onto Y. Existing methods pose it as a constrained least-squares problem in terms of $R$,
\vspace{-1.5mm} 
\begin{equation}
\mathcal{F}(R) {=} \dfrac{1}{K} \sum_{n=1}^{K} \norm{y_n - Rx_n}^2\text{, subject to } R^{*}R{=}RR^{*}{=}I_d, 
\label{eq:general_cost_function}
\vspace{-1.5mm}
\end{equation}  
in which $^*$ denotes the conjugate transpose, and $I_d$ is the identity matrix. To minimize~\eqref{eq:general_cost_function}, some methods in the literature estimate $R$ directly~\cite{Umeyama1991} by calculating the PCDs cross-covariance matrix and performing a singular value decomposition (SVD)~\cite{Horn87,JacobiSVD}. Others use quaternion algebra to represent rotations, recovering the equivalent matrix via a well-known relation~\cite{Walker1991,Besl1992,Zhang1994}.  

To estimate a (transformation) matrix, one may consider using Kronecker products and vectorization~\cite{Sayed08}. However, the matrix size and the possible constraints to which its entries are subject might result in extensive analytic procedures and expendable computational complexity.

Describing 3D rotations via quaternions has several advantages over matrices, e.g., intuitive geometric interpretation, and independence of the coordinate system~\cite{Girard2007}. Particularly, quaternions require only one constraint -- the rotation quaternion should have norm equal to one -- whereas rotation matrices need six: each row must be a unity vector (norm one) and the columns must be mutually orthogonal (see~\cite{quaternionReport}, p.30). Nevertheless, performing standard vector calculus in quaternion algebra (to calculate the gradient of the error vector) incur a cumbersome analytic derivation~\cite{Mandic2011,Jahanchahi2012,Mengdi2014}. To circumvent that, \eqref{eq:general_cost_function} is recast in GA (which encompasses quaternion algebra) by introducing the concept of \emph{multivectors}. This allows for utilizing GC to obtain a neat and compact analytic derivation of the gradient of the error vector. Using that, the GA-based AF is conceived without restrictions to the dimension of the underlying vector space (otherwise impossible with quaternion algebra), allowing it to be readily applicable to high-dimensional ($\mathbb{R}^n, n>3$) rotation estimation problems (\hspace{-0.1mm}\cite{hestenes1999newfoundations}, p.581).%

\section{Geometric-Algebra Approach}
\label{sec:GA_approach}

\subsection{Elements of Geometric Algebra}
\label{ssec:elements_of_GA}
In a nutshell, the GA $\mathcal{G}(\mathbb{R}^n)$ is a geometric extension of $\mathbb{R}^n$ which enables algebraic representation of orientation and magnitude. Vectors in $\mathbb{R}^n$ are also vectors in $\mathcal{G}(\mathbb{R}^n)$. Each orthogonal basis in $\mathbb{R}^n$, together with the scalar $1$, generates $2^n$ members (\emph{multivectors}) of $\mathcal{G}(\mathbb{R}^n)$ via the \textit{geometric product} operated over the $\mathbb{R}^n$ (\hspace{-0.1mm}\cite{hestenes1987GAcalculus}, p.19).

Consider vectors $a$ and $b$ in $\mathbb{R}^n$. The geometric product is defined as $ab \triangleq a \cdot b + a \wedge b$, in terms of the inner ($\cdot$) and outer ($\wedge$) products (\hspace{-0.1mm}\cite{Hitzer_Introduction}, Sec. $2.2$). Note that in general the geometric product is noncommutative because $a \wedge b = -(b \wedge a)$. In this text, from now on, all products are geometric products.

\begin{figure}[!t]
\centering
\includegraphics[height=0.15\textwidth]{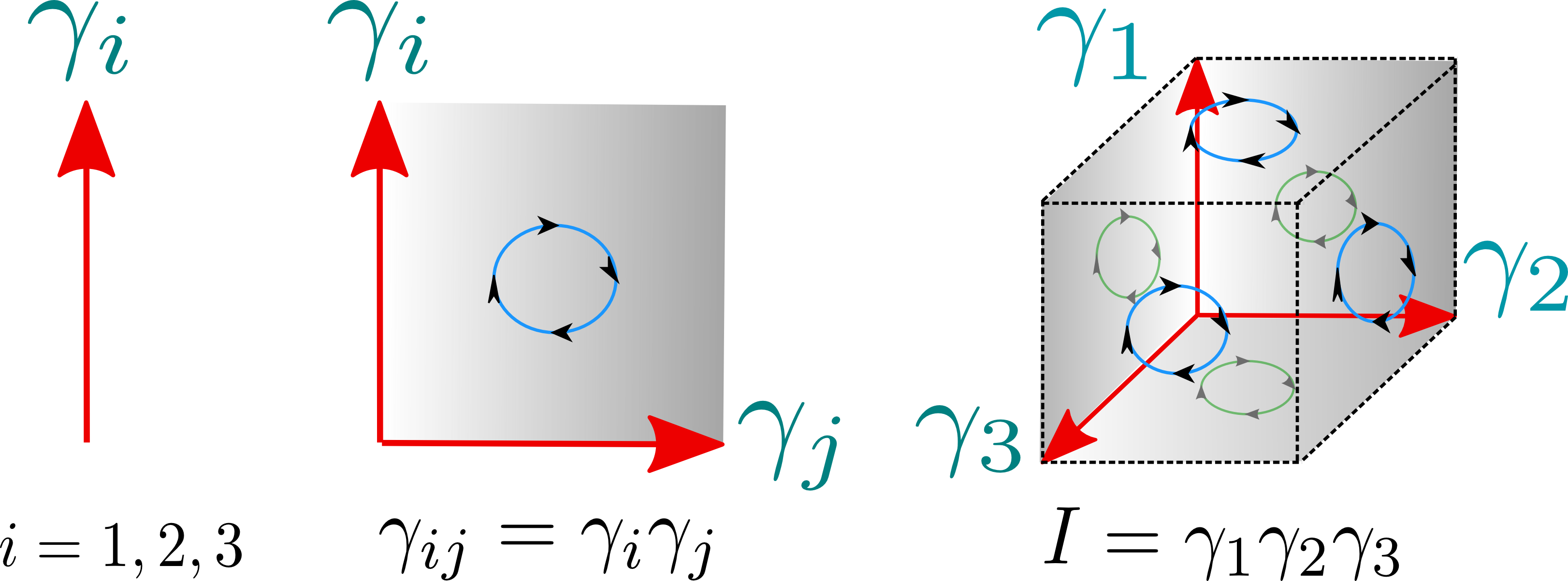}
\caption{The elements of $\mathcal{G}(\mathbb{R}^3)$ basis (besides the scalar $1$): $3$ vectors, $3$ bivectors (oriented areas) $\gamma_{ij}$, and the pseudoscalar $I$ (trivector/oriented volume).}
\label{fig:multivectors_in_R3}
\end{figure}

For the $\mathbb{R}^3$ case, $\mathcal{G}(\mathbb{R}^3)$ has dimension $2^3 = 8$, with basis $\{1,\gamma_1,\gamma_2,\gamma_3,\gamma_{12},\gamma_{23},\gamma_{31},I\}$, i.e., one scalar, three orthogonal vectors $\gamma_{i}$ (basis for $\mathbb{R}^3$), three bivectors $\gamma_{ij} \triangleq \gamma_{i}\gamma_{j}=\gamma_{i}\wedge\gamma_{j}, i \neq j$ ($\gamma_{i}\cdot\gamma_{j} = 0, i \neq j$), and one trivector (\textit{pseudoscalar}) $I \triangleq \gamma_1\gamma_2\gamma_3$ (\figurename\ref{fig:multivectors_in_R3}). To illustrate the geometric multiplication, take two vectors $a = \gamma_1$ and $b = 2\gamma_1 + 4\gamma_3$. Then, $ab = \gamma_1(2\gamma_1 + 4\gamma_3) = \gamma_1\cdot(2\gamma_1 + 4\gamma_3) + \gamma_1\wedge(2\gamma_1 + 4\gamma_3) = 2 + 4(\gamma_1\wedge\gamma_3) = 2 + 4\gamma_{13}$ (a scalar plus a bivector). 

The basic element of a GA is a multivector $A$,
\vspace{-1.5mm}
\begin{equation}
A = \langle A \rangle_0 + \langle A \rangle_1 + \langle A \rangle_2 + \cdots = \sum_{g} \langle A \rangle_g \text{,} 
\vspace{-2mm}
\end{equation} 
which is comprised of its g-grades (or g-vectors) $\langle\cdot\rangle_g$, e.g., $g=0$ (scalars), $g=1$ (vectors), $g=2$ (bivectors), $g=3$ (trivectors), and so on. The ability to group together scalars, vectors, and hyperplanes in an unique element (the multivector $A$) is the foundation on top of which GA theory is built on. Except where otherwise noted, scalars ($g = 0$) and vectors ($g = 1$) are represented by lower-case letters, e.g., $a$ and $b$, and general multivectors by upper-case letters, e.g., $A$ and $B$. Also, in $\mathbb{R}^3$, $\langle A \rangle_g = 0$, $g > 3$ (\hspace{-0.1mm}\cite{hestenes1999newfoundations}, p.42).
  
The \textit{reverse} of a multivector $A$ (analogous to conjugation of complex numbers and quaternions) is defined as
\vspace{-2mm}
\begin{equation}
\widetilde{A} \triangleq \sum_{g=0}^{n} (-1)^{g(g-1)/2}\langle A \rangle_g\text{.}
\label{eq:reversion}
\vspace{-2mm}
\end{equation}
For example, the reverse of the bivector $A {=} \langle A \rangle_0 {+} \langle A \rangle_1 {+} \langle A \rangle_2$ is $\widetilde{A} {=} \widetilde{\langle A \rangle_0} {+} \widetilde{\langle A \rangle_1} {+} \widetilde{\langle A \rangle_2} {=} A_0 {+} A_1 {-} A_2$. 

The GA \textit{scalar product} $*$ between two multivectors $A$ and $B$ is defined as $A {*} B = \langle AB \rangle$, in which $\langle \cdot \rangle \equiv \langle \cdot \rangle_0$. From that, the \emph{magnitude} of a multivector is defined as $|A|^2 {=} A {*} \~A {=} \sum_{g} |A|_g^2$.

\subsection{The Estimation Problem in GA}
\label{ssec:estimation_in_GA}
The problem~\eqref{eq:general_cost_function} may be posed in GA as follows. The rotation matrix $R$ in the error vector $y_n - Rx_n$ is substituted by the rotation operator comprised by the bivector $r$ (the only bivector in this paper written with lower-case letter) and its \textit{reversed} version $\widetilde{r}$~\cite{hestenes1999newfoundations},
\vspace{-1.5mm} 
\begin{equation}
e_n = y_n - r x_n \widetilde{r} \text{, subject to } r\~r = \~rr = |r|^2 = 1.
\label{eq:error_GAAF_3D}
\vspace{-1.5mm}
\end{equation}
Thus, $r$ is a \textit{unit rotor} in $\mathcal{G}(\mathbb{R}^3)$. Note that the term $r x_n \widetilde{r}$ is simply a rotated version of the vector $x_n$ (\hspace{-0.1mm}\cite{Hitzer_Introduction}, Eq.54). This rotation description is similar to the one provided by quaternion algebra. In fact, it can be shown that the subalgebra of $\mathcal{G}(\mathbb{R}^3)$ containing only the multivectors with even grades (rotors) is \textit{isomorphic} to quaternions~\cite{Hitzer_Introduction}. However, unlike quaternions, GA enables to describe rotations in any dimension. More importantly, with the support of GC, optimization problems can be carried out in a clear and compact manner~\cite{Hitzer_MultivectorDiffCalc,hestenes1987GAcalculus}. 

Hypercomplex AFs available in the literature make use of quaternion algebra~\cite{Took_QLMS,Neto_WLQLMS_2011} and even GA theory~\cite{Hitzer_split}. However, the error vector therein has the form $e = y - r x$, which is not appropriate to model rotation error since it lacks $\widetilde{r}$ multiplying $x$ from the right.

This way, \eqref{eq:general_cost_function} is rewritten using~\eqref{eq:error_GAAF_3D}, generating a new cost function,
\vspace{-2.5mm} 
\begin{equation}
J(r) = \dfrac{1}{K}\sum_{n=1}^{K} e_n*\widetilde{e}_n = \dfrac{1}{K}\sum_{n=1}^{K} \langle e_n\widetilde{e}_n\rangle = \dfrac{1}{K}\sum_{n=1}^{K}{|e_n|}^2 \text{,}
\label{eq:new_cost_function}
\vspace{-1.5mm}
\end{equation}
subject to $r\~r = \~rr = |r|^2 = 1$.

\section{Geometric-Algebra LMS}
\label{sec:GA-LMS}

The GA-LMS designed in the sequel should make $r x_n \widetilde{r}$ as close as possible to $y_n$ in order to minimize~\eqref{eq:new_cost_function}. The AF provides an estimate for the bivector $r$ via a recursive rule of the form,
\vspace{-1.5mm}
\begin{equation}
r_i = r_{i-1} + \mu G \text{,}
\label{eq:GAAF_update}
\vspace{-1.5mm}
\end{equation}  
where $i$ is the (time) iteration, $\mu$ is the AF step size, and $G$ is a multivector-valued update quantity related to the estimation error~\eqref{eq:error_GAAF_3D} (analogous to the standard formulation in~\cite{Sayed08}, p.143).

A proper selection of $G$ is required to enforce $J(r_i) < J(r_{i-1})$ at each iteration. This work adopts the steepest-descent rule~\cite{Sayed08,Diniz2013}, in which the AF is designed to follow the opposite direction of the reversed gradient of the cost function, namely $\widetilde{\nabla} J(r_{i-1})$ (note the analogy between the reversed $\widetilde{\nabla}$ and the hermitian conjugate $\nabla^*$ from the standard formulation). This way, $G$ is proportional to $\widetilde{\nabla} J(r_{i-1})$, 
\vspace{-1.5mm}
\begin{equation}
G \triangleq -B\widetilde{\nabla} J(r_{i-1})\text{,}
\label{eq:definition_P}
\vspace{-1.5mm}
\end{equation}
in which $B$ is a general multivector, in contrast with the standard case in which $B$ would be a matrix~\cite{Sayed08}. In the AF literature, setting $B$ equal to the identity matrix results in the steepest-descent update rule (\hspace{-0.1mm}\cite{Sayed08}, Eq. 8-19). In GA though, the multiplicative identity is the multivector (scalar) $1$ (\hspace{-0.1mm}\cite{hestenes1987GAcalculus}, p.3), thus $B=1$.   
     
Embedding $1/K$ into $J(r)$ and expanding yields,
\vspace{-1.5mm}  
\begin{equation}
\begin{array}{ll}
J(r) \hspace{-2mm} &{=} \sum\limits_{n=1}^{K} \big(y_n {-} r x_n \widetilde{r}\big){*}\big(y_n {-} r x_n \widetilde{r}\widetilde{\big)}\\ 
       &{=} \sum\limits_{n=1}^{K} \left[y_n*\widetilde{y}_n {-} y_n {*} (r \widetilde{x} \widetilde{r}) {-} (r x \widetilde{r}) {*} \widetilde{y}_n 
       {+} (r x \widetilde{r}){*}(r x \widetilde{r}\widetilde{)}\right]\\       
       &{=} \sum\limits_{n=1}^{K} |y_n|^2 {+} |x_n|^2 {-} 2\langle y_nrx_n\widetilde{r}\rangle \text{,} 
\end{array}
\label{eq:expand_J}
\end{equation}
where the reversion rule~\eqref{eq:reversion} was used to conclude that $y_n = \widetilde{y}_n$, $x_n = \widetilde{x}_n$ (they are vectors), and $r\widetilde{r} = \widetilde{r}r = 1$.

Using Geometric calculus techniques~\cite{hestenes1987GAcalculus,Hitzer_MultivectorDiffCalc,Lasenby1998}, the gradient of $J(r)$ is calculated from~\eqref{eq:expand_J}, 
\vspace{-2mm}
\begin{equation}
\begin{array}{ll}
\nabla J(r) = \partial_r J(r) \hspace{-2mm}&= -2\partial_r\sum\limits_{n=1}^{K} \langle y_nrx_n\widetilde{r}\rangle \\
&= -2 \left[ \sum\limits_{n=1}^{K} \partial_r\langle \dot{r} M_n \rangle + \partial_r \langle T_n \dot{\widetilde{r}}\rangle\right] \text{,} 
\end{array}
\label{eq:gradient_J}
\end{equation}
in which the product rule (\hspace{-0.1mm}\cite{Hitzer_MultivectorDiffCalc}, Eq.~5.12) was used and the overdots emphasize which quantity is being differentiated by $\partial_r$ (\hspace{-0.1mm}\cite{Hitzer_MultivectorDiffCalc}, Eq. 2.43). The terms $M_n = x_n\widetilde{r}y_n$ and $T_n = y_n r x_n$ are obtained applying the \textit{cyclic reordering property} $\langle AD \cdots C \rangle = \langle D \cdots CA \rangle = \langle CAD \cdots \rangle$~\cite{Hitzer_Introduction}. The first term on the right-hand side of~\eqref{eq:gradient_J} is $\partial_r\langle \dot{r} M_n \rangle {=} M_n$~(\hspace{-0.1mm}\cite{Hitzer_MultivectorDiffCalc}, Eq.~7.10), and the second term is $\partial_r \langle T_n \dot{\widetilde{r}}\rangle {=} {-}\widetilde{r}T_n\widetilde{r} {=} {-}\widetilde{r}(y_n r x_n)\widetilde{r}$ (see the Appendix). Plugging back into~\eqref{eq:gradient_J}, the GA-form of the gradient of $J(r)$ is obtained 
\vspace{-2.5mm}
\begin{equation}
\begin{array}{ll}
\partial_r J(r)\hspace{-3mm}&{=} {-}2 \sum\limits_{n=1}^{K} x_n\widetilde{r}y_n  {-}  \widetilde{r}(y_n r x_n)\widetilde{r} \\
&{=} {-}2 \widetilde{r}\sum\limits_{n=1}^{K} (rx_n\widetilde{r})y_n  {-}  y_n (r x_n\widetilde{r}){=} 4 \widetilde{r}\sum\limits_{n=1}^{K} y_n \wedge (r x_n\widetilde{r})\text{,}
\end{array}
\label{eq:gradient_J_2}
\end{equation}
where the relation $ab - ba = 2(a \wedge b)$ was used~(\hspace{-0.1mm}\cite{hestenes1999newfoundations}, p.39). 

In~\cite{Lasenby1998}, the GA framework to handle linear transformations is applied for mapping~\eqref{eq:gradient_J_2} back into matrix algebra, obtaining a rotation matrix (and not a rotor). Here, on the other hand, the algorithm steps are completely carried out in GA (design and computation), since the goal is to devise an AF to estimate a multivector quantity (rotor) for PCDs rotation problems.     

Substituting~\eqref{eq:gradient_J_2} into~\eqref{eq:definition_P} (with $B=1$, as aforementioned) and explicitly showing the term $1/K$ results in
\vspace{-1.5mm}
\begin{equation}
G = \dfrac{4}{K}\left[\sum\limits_{n=1}^{K} y_n \wedge (r_{i-1} x_n\widetilde{r}_{i-1})\right]r_{i-1},
\label{eq:G}
\end{equation}
which upon plugging into~\eqref{eq:GAAF_update} yields
\vspace{-1.5mm}
\begin{equation}
\boxed{r_i = r_{i-1} + \mu{\dfrac{4}{m}}\left[\sum\limits_{n=1}^{m} y_n \wedge (r_{i-1} x_n\widetilde{r}_{i-1})\right]r_{i-1}} \text{,}
\label{eq:steepest_descent}
\vspace{-1.5mm}
\end{equation}  
where a substitution of variables was performed to enable writing the algorithm in terms of a rank captured by $m$, i.e., one can select $m \in [1,K]$ to choose how many correspondence pairs are used at each iteration. This allows for balancing computational cost and performance, similar to the Affine Projection Algorithm (APA) rank~\cite{Diniz2013,Sayed08}. If $m=K$, \eqref{eq:steepest_descent} uses all the available points, originating the \emph{geometric-algebra steepest-descent} algorithm. This paper focuses on the case $m=1$ (one pair per iteration) which is equivalent to approximating $\widetilde{\nabla} J(r)$ by its \textit{current value} in~\eqref{eq:G}~\cite{Sayed08}, 
\vspace{-1.5mm}
\begin{equation}
{\dfrac{4}{K}}\left[\sum\limits_{n=1}^{K} y_n \wedge (r_{i-1} x_n\widetilde{r}_{i-1})\right]r_{i-1} {\approx} 4\left[y_i \wedge (r_{i-1} x_i\widetilde{r}_{i-1})\right]r_{i-1}\text{,} 
\label{eq:instantaneous_approx_grad_J}
\end{equation}
resulting in the GA-LMS update rule,
\begin{equation}
\boxed{r_i = r_{i-1} + \mu\left[y_i \wedge (r_{i-1} x_i\widetilde{r}_{i-1})\right]r_{i-1}} \text{,}
\label{eq:GA-LMS}
\end{equation}
in which the factor $4$ was absorbed by $\mu$. Note that~\eqref{eq:GA-LMS} was obtained without restrictions to the dimension of the vector space containing $\{y_n,x_n\}$. 

Adopting~\eqref{eq:instantaneous_approx_grad_J} has an important practical consequence for the registration of PCDs. Instead of ``looking at'' the sum of all correspondence-pairs outer products ($m=K$), when $m=1$ the filter uses only the pair at iteration $i$, $\{y_i,x_i\}$, to update $r_{i-1}$. Thus, from an information-theoretic point of view, the GA-LMS uses \emph{less information per iteration} when compared to methods in the literature~\cite{JacobiSVD,Horn87,Walker1991,Umeyama1991,Besl1992,Zhang1994} that require all the correspondences at each algorithm iteration.

From GA theory it is known that any multiple of a unit rotor $q$, namely $\lambda q, \lambda \in \mathbb{R} \setminus \{0\}$, $|\lambda q| = \lambda$, provides the same rotation as $q$. However, it scales the magnitude of the rotated vector by a factor of $\lambda^2$, $|(\lambda q)x\widetilde{(\lambda q)}|=\lambda^2|x|$. Thus, to comply with $r\~r = \~rr = |r_i|^2 = 1$ (see~\eqref{eq:new_cost_function}) and avoid scaling the PCD points, the estimate $r_i$ in~\eqref{eq:GA-LMS} is normalized at each iteration when implementing the GA-LMS.%

\emph{Note on computational complexity.} The computational cost is calculated by breaking~\eqref{eq:GA-LMS} into parts. The term $r_{i-1} x_i\widetilde{r}_{i-1}$ has two geometric multiplications, which amounts to $28$ real multiplications (RM) and $20$ real additions (RA). The outer product $y_i \wedge (r_{i-1} x_i\widetilde{r}_{i-1})$ amounts to $6$ RM and $3$ RA. The evaluation of $\mu\left[y_i \wedge (r_{i-1} x_i\widetilde{r}_{i-1})\right]r_{i-1}$ requires more $20$ RM and $12$ RA. Finally, $r_{i-1} + \mu\left[y_i \wedge (r_{i-1} x_i\widetilde{r}_{i-1})\right]r_{i-1}$ requires more $4$ RA. Summarizing, the cost of the GA-LMS is $54$ RM and $39$ RA per iteration. SVD-based methods compute the covariance matrix of the $3\times K$ PCDs at each iteration, which has the cost $O(K)$, i.e., it depends on the number of points. This suggests that adopting the GA-LMS instead of SVD can contribute to reduce the computational cost when registering PCDs with a great number of points, particularly when $K \gg 54$.

\section{Simulations} 
\label{sec:sims}

Given $K$ corresponding source and target points (X and Y), the GA-LMS estimates the rotor $r$ which aligns the input vectors in X to the desired output vectors in Y. At first, a ``toy problem'' is provided depicting the alignment of two cubes PCDs. Then, the AF performance is further tested when registering two PCDs from the ``Stanford Bunny''\footnote{This paper has supplementary downloadable material available at www.lps.usp.br/wilder, provided by the authors. This includes an .avi video showing the alignment of the PCD sets, the MATLAB code to reproduce the simulations, and a readme file. This material is 25 MB in size.}, one of the most popular 3D datasets~\cite{Turk_Zippered_1994}. 

The GA-LMS is implemented using the GAALET C++ library~\cite{seybold_gaalet_2010} which enables users to compute the geometric product (and also the outer and inner products) between two multivectors. For all simulations, the rotor initial value is $r = 0.5 + 0.5\gamma_{12} + 0.5\gamma_{23} + 0.5\gamma_{31}$ ($|r| = 1$). 

\subsection{Cube registration}
\label{ssec:cube}

\begin{figure}[!t]
\centering
\includegraphics[width=0.48\textwidth]{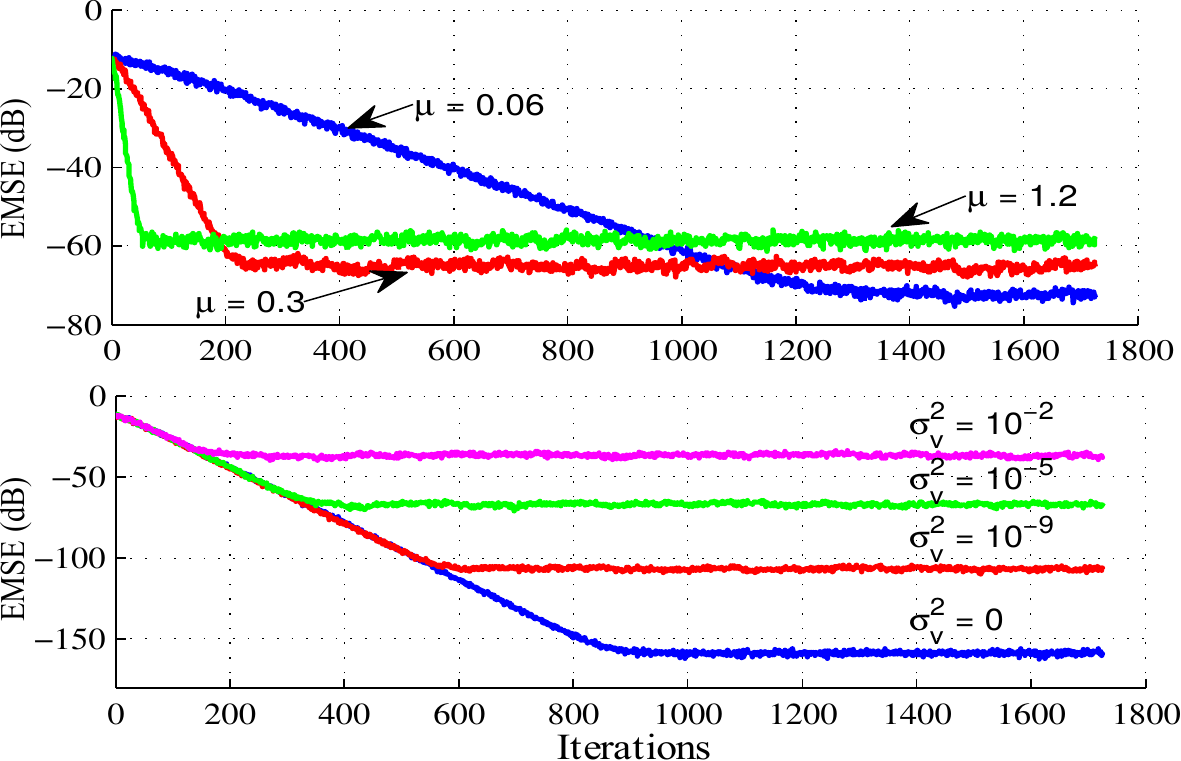}
\caption{Cube set. (top) EMSE for $\sigma_v^2 = 10^{-5}$ and different values of $\mu$. (bottom) EMSE for $\mu=0.2$ and different noise variances $\sigma_v^2$. For all cases, the steady state is achieved using only part of the correspondence points. The curves are averaged over $200$ realizations.}
\label{fig:cube_curves}
\end{figure}

Two artificial cube PCDs with edges of $0.5$ meters and $K=1728$ points were created. The relative rotation between the source and target PCDs is $120^{\circ}$, $90^{\circ}$, and $45^{\circ}$, about the $x,y,$ and $z$ axes, respectively. Simulations are performed assuming different levels of measurement noise in the points of the Target PCD, i.e., $y_i$ is perturbed by $v_i$, a $3{\times}1$ random vector with entries drawn from a white Gaussian process of variance $\sigma^2_v \in \{0, 10^{-9}, 10^{-5}, 10^{-2}\}$.  

\figurename~\ref{fig:cube_curves} shows curves of the excess mean-square error (EMSE$(i)=\expec|y_i - r_{i-1}x_i\widetilde{r}_{i-1}|^2$) averaged over 200 realizations. \figurename~\ref{fig:cube_curves} (top) depicts the typical trade-off between convergence speed and steady-state error when selecting the values of $\mu$ for a given $\sigma^2_v$, e.g., for $\mu=0.3$ the filter takes around $300$ iterations (correspondence pairs) to converge, whereas for $\mu=0.06$ it needs around $1400$ pairs. \figurename~\ref{fig:cube_curves} (bottom) shows how the AF performance is degraded when $\sigma^2_v$ increases. The correct rotation is recovered for all cases above. For $\sigma^2_v > 10^{-2}$ the rotation error approaches the order of magnitude of the cube edges (0.5 meters). For the noise variances in \figurename~\ref{fig:cube_curves} (bottom), the SVD-based method~\cite{JacobiSVD} implemented by the Point Cloud Library (PCL)~\cite{Rusu_ICRA2011_PCL} achieves similar results except for $\sigma^2_v = 0$, when SVD reaches $-128dB$ compared to $-158dB$ of GA-LMS.       
 
\subsection{Bunny registration}
\label{ssec:bunny}       

\begin{figure}[!t]
\centering
\subfloat[]{\includegraphics[width=0.15\textwidth]{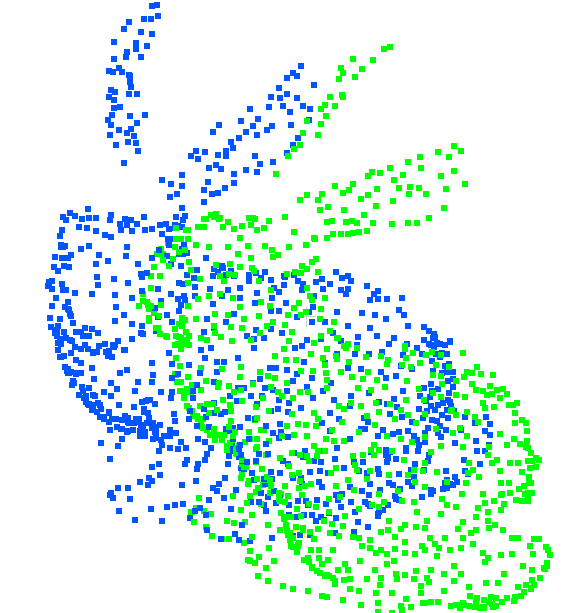}
\label{fig:bunny_PCs_unaligned}}
\hfil
\vspace{-1mm}
\subfloat[]{\includegraphics[width=0.15\textwidth]{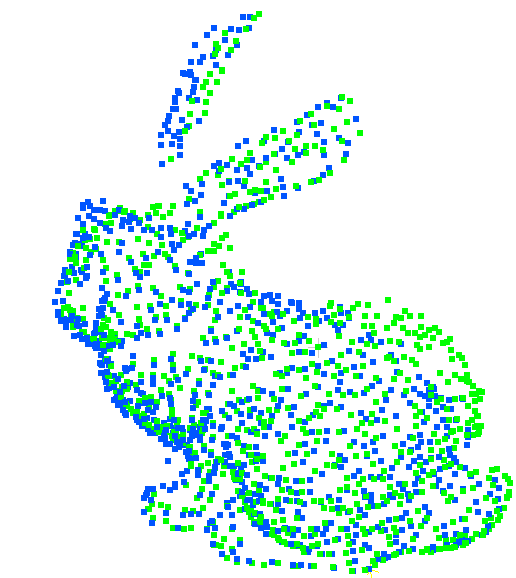}
\label{fig:bunny_PCs_aligned}}
\caption{PCDs of the bunny set. (a) Unaligned, (b) after GA-LMS alignment.}
\label{fig:bunny_PCs}
\end{figure}

Two specific scans of the ``Stanford Bunny'' dataset~\cite{Turk_Zippered_1994} are selected (see~\figurename~\ref{fig:bunny_PCs}), with a relative rotation of $45^{\circ}$ about the $z$ axis. Each bunny has an average nearest-neighbor (NN) distance of around $0.5mm$. The correspondence between source and target points is pre-established using the matching system described in~\cite{alnuaimi_indoor_2015}. It suffices to say the point matching is not perfect and hence the number of true correspondence (TCs) and its ratio with respect to the total number of correspondences is $191/245 = 77\%$. 

\begin{figure}[!t]
\centering
\includegraphics[width=0.5\textwidth]{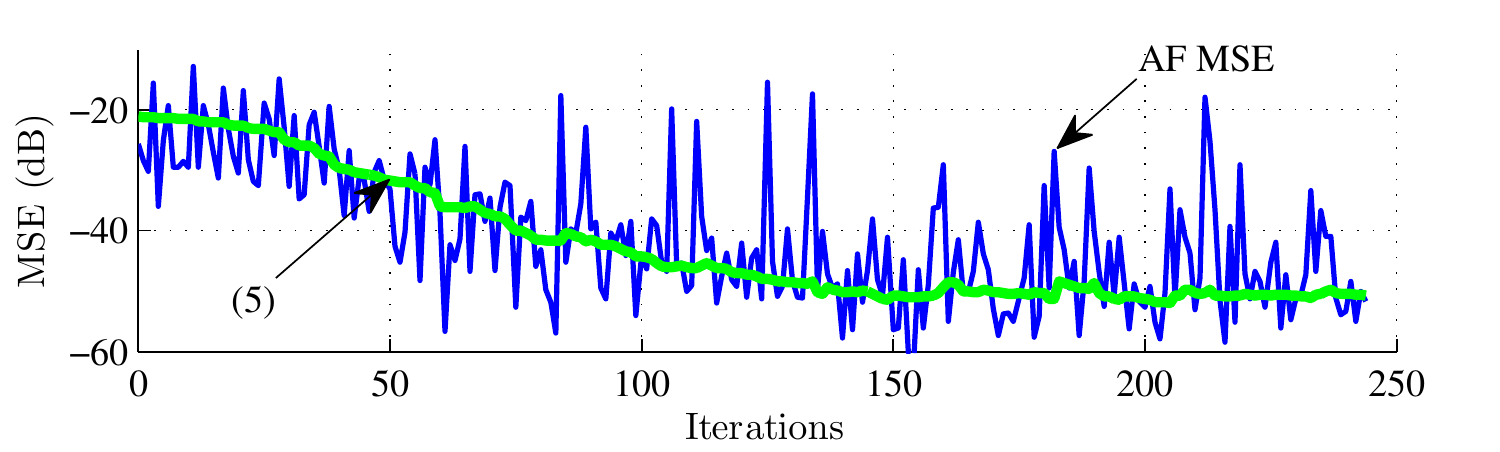}
\caption{Bunny set, $\mu=8$. The cost function (5) curve is plotted on top of the MSE to emphasize the minimization performed by the AF. The steady state is reached before using all the available correspondences.}
\label{fig:bunny_curves}
\end{figure}

The performance of the GA-LMS with $\mu=8$ (selected via extensive parametric simulations) is depicted in \figurename~\ref{fig:bunny_curves}. It shows the curve (in blue) for the mean-square error (MSE), which is approximated by the instantaneous squared error (MSE$(i)\approx|d_i - r_{i-1}x_i\widetilde{r}_{i-1}|^2$), where $d_i = y_i + v_i$ is the noise-corrupted version of $y_i$ (in order to model acquisition noise in the scan). As in a real-time online registration, the AF runs only one realization, producing a noisy MSE curve (it is not an ensemble average). Nevertheless, from the cost function~\eqref{eq:new_cost_function} curve (in green), plotted on top of the MSE using only the good correspondences, one can see the GA-LMS minimizes it, achieving a steady-state error of $-50.67dB$ at $i \approx 210$. The PCL SVD-based method achieves a slightly lower error of $-51.81dB$ (see supplementary material), although using all the $245$ pairs at each iteration. The GA-LMS uses only 1 pair. 

\section{Conclusion}
\label{sec:conclusion}
This work introduced a new AF completely derived using GA theory. The GA-LMS was shown to be functional when estimating the relative rotation between two PCDs, achieving errors similar to those provided by the SVD-based method used for comparison. The GA-LMS, unlike the SVD method, allows to assess each correspondence pair individually (one pair per iteration). That fact is reflected on the GA-LMS computational cost per iteration -- it does not depend on the number of correspondence pairs (points) $K$ to be processed, which can lower the computation cost (compared to SVD) of a complete registration algorithm, particularly when $K$ grows. To improve performance, some strategies could be adopted: reprocessing iterations in which the MSE$(i)$ changes abruptly, and data reuse techniques~\cite{Wilder11,Wilder13,Chamon14}. A natural extension is to generalize the method to estimate multivectors of any grade, covering a wider range of applications.

\appendix
For a general multivector $A$ and a unit rotor $\Omega$, it holds that
\vspace{-2mm}
\begin{equation}
\partial_{\Omega}\langle A \dot{\widetilde{\Omega}} \rangle = - \widetilde{\Omega} A \widetilde{\Omega} \text{.} 
\vspace{-2mm} 
\label{prop:dorans_result}
\end{equation}
\begin{IEEEproof}
Given that the scalar part (0-grade) of a multivector is not affected by rotation $(\partial_{\Omega}\langle \Omega A \widetilde{\Omega} \rangle = 0)$, and using the product rule, one can write $\partial_{\Omega}\langle \Omega A \widetilde{\Omega} \rangle = A \widetilde{\Omega} + \partial_{\Omega} \langle \Omega A \dot{\widetilde{\Omega}} \rangle = 0 \text{,}$
\vspace{-2mm}
\begin{equation}
\partial_{\Omega} \langle \Omega A \dot{\widetilde{\Omega}} \rangle = -A \widetilde{\Omega}\text{.}
\vspace{-2mm}
\label{eq:dorans_aux_1}
\end{equation}
Using the scalar product definition, the cyclic reordering property, and Eq. (7.2) in~\cite{Hitzer_MultivectorDiffCalc}, $\partial_{\Omega} \langle \Omega A \dot{\widetilde{\Omega}} \rangle {=} \partial_{\Omega} \left[ \dot{\widetilde{\Omega}} {*} (\Omega A) \right] {=} \\ \left[(\Omega A) {*} \partial_{\Omega}\right] \widetilde{\Omega}.$ Plugging back into~\eqref{eq:dorans_aux_1} and multiplying by $\widetilde{\Omega}$ from the left, $\widetilde{\Omega}\left[(\Omega A) * \partial_{\Omega}\right] \widetilde{\Omega} {=} {-}\widetilde{\Omega}A \widetilde{\Omega}$. Since the term $\left[(\Omega A) * \partial_{\Omega}\right]$ is an algebraic scalar, $\widetilde{\Omega}\left[(\Omega A) * \partial_{\Omega}\right] \widetilde{\Omega} {=} \left[(\Omega A) * \partial_{\Omega}\right] \widetilde{\Omega} \widetilde{\Omega} {=} \partial_{\Omega}\left[\dot{(\widetilde{\Omega}\widetilde{\Omega})}*(\Omega A)\right] {=} \\ \partial_{\Omega}\langle\dot{(\widetilde{\Omega}\widetilde{\Omega})}\Omega A\rangle {=} \partial_{\Omega}\langle A \dot{\widetilde{\Omega}}\rangle \Rightarrow \boxed{\partial_{\Omega}\langle A \dot{\widetilde{\Omega}}\rangle = - \widetilde{\Omega} A \widetilde{\Omega}}$.  
\end{IEEEproof}

\ifCLASSOPTIONcaptionsoff
  \newpage
\fi



%\bibliographystyle{IEEEtran}
\bibliographystyle{IEEEbib.bst}
\bibliography{refs}

\end{document}